\documentclass[letterpaper, 10 pt, conference]{ieeeconf}  

\IEEEoverridecommandlockouts                              

\title{\LARGE \bf
Geometry-Aware Unsupervised Domain Adaptation for Stereo Matching}

\author{Hiroki Sakuma$^{1}$ and Yoshinori Konishi$^{1}$
\thanks{$^{1}$SenseTime Japan Ltd., {\tt\small \{sakuma,konishi\}@sensetime.jp}}%
}

\newcommand{\argmin}{\mathop{\rm argmin}\limits}
\newcommand{\softmax}{\mathop{\rm softmax}\limits}

\usepackage{here}
\usepackage{color}
\usepackage{dsfont}
\usepackage{comment}
\usepackage{amsmath}
\usepackage{cases}
\usepackage{amsfonts}
\usepackage{multirow}
\usepackage{graphicx}
\usepackage{indentfirst}
\usepackage[labelformat=empty]{subcaption}

\begin{document}

\maketitle
\thispagestyle{empty}
\pagestyle{empty}

\begin{abstract}

Recently proposed DNN-based stereo matching methods that learn priors directly from data are known to suffer a drastic drop in accuracy in new environments. 
Although supervised approaches with ground truth disparity maps often work well, collecting them in each deployment environment is cumbersome and costly. 
For this reason, many unsupervised domain adaptation methods based on image-to-image translation have been proposed, but these methods do not preserve the geometric structure of a stereo image pair because the image-to-image translation is applied to each view separately. 
To address this problem, in this paper, we propose an attention mechanism that aggregates features in the left and right views, called Stereoscopic Cross Attention (SCA). 
Incorporating SCA to an image-to-image translation network makes it possible to preserve the geometric structure of a stereo image pair in the process of the image-to-image translation. 
We empirically demonstrate the effectiveness of the proposed unsupervised domain adaptation based on the image-to-image translation with SCA.

\end{abstract}

\section{Introduction}

\begin{figure}[t]
    \centering
        \begin{tabular}{ccc}
          \begin{minipage}{0.01\hsize}
            \centering
                \subcaption{(a)}
          \end{minipage} &
          \begin{minipage}{0.40\hsize}
            \centering
                \includegraphics[width=3.8cm]{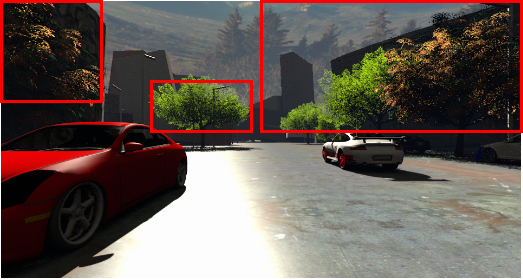}
          \end{minipage} &
          \begin{minipage}{0.40\hsize}
            \centering
                \includegraphics[width=3.8cm]{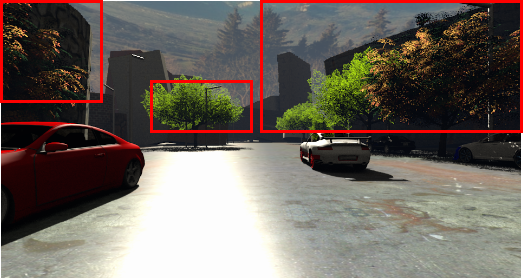}
          \end{minipage} \\
          \begin{minipage}{0.01\hsize}
            \centering
                \subcaption{(b)}
          \end{minipage} &
          \begin{minipage}{0.40\hsize}
            \centering
                \includegraphics[width=3.8cm]{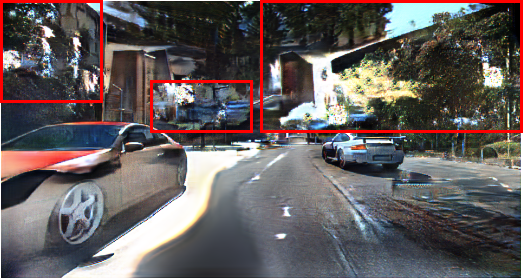}
          \end{minipage} &
          \begin{minipage}{0.40\hsize}
            \centering
                \includegraphics[width=3.8cm]{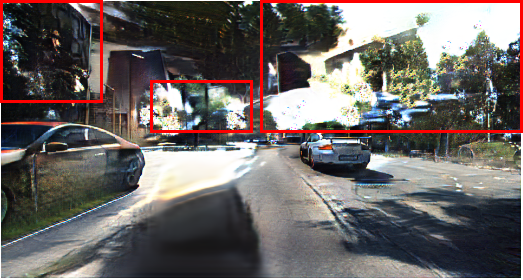}
          \end{minipage} \\
          \begin{minipage}{0.01\hsize}
            \centering
                \subcaption{(c)}
                \subcaption{}
          \end{minipage} &
          \begin{minipage}{0.40\hsize}
            \centering
                \includegraphics[width=3.8cm]{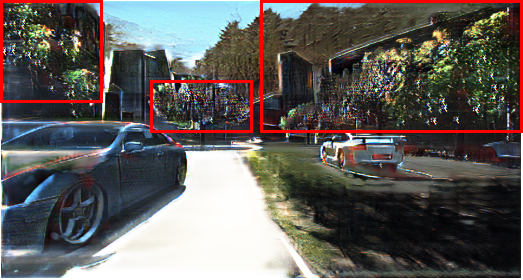}
                \subcaption{Left image}
          \end{minipage} &
          \begin{minipage}{0.40\hsize}
            \centering
                \includegraphics[width=3.8cm]{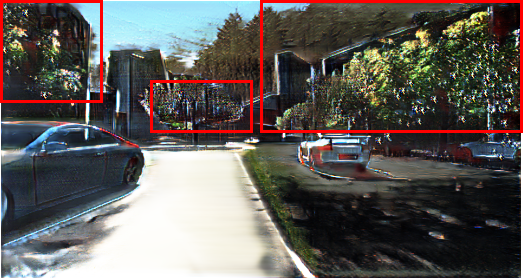}
                \subcaption{Right image}
          \end{minipage}
        \end{tabular}
        \caption{The image-to-image translation results from Driving to KITTI Stereo 2015. (a) No translation, (b) separately translated, (c) translated with the proposed SCA.}
        \label{fig:translation_results}
\end{figure}

Stereo matching is used for depth prediction from a rectified stereo image pair and it plays an essential role in many computer vision applications such as autonomous driving and robot navigation. 
It has been widely investigated for several decades and a vast number of approaches based on local \cite{adaptive_support}, global \cite{graph_cut} or semi-global \cite{sgm} optimization have been proposed. 
In recent years, convolutional neural networks (CNNs) have been successfully applied to stereo matching in an end-to-end manner \cite{scene_flow,psmnet,gwcnet} and they have achieved higher accuracy than the traditional approaches in challenging real-world scenarios, like KITTI benchmarks \cite{kitti2012,kitti2015}. 
Although CNN-based approaches are very accurate, they are known to suffer a drastic drop in accuracy in new environments. 
This problem is usually addressed by fine-tuning on a small real dataset \cite{kitti2012,kitti2015} after training on a large synthetic dataset \cite{scene_flow,virtual_kitti}. 
Collecting even such a small amount of real data with the ground truth disparity maps in each deployment environment is, however, cumbersome and costly because it requires an expensive depth sensor and careful calibration between the depth sensor and the stereo cameras. 
To address this problem, many unsupervised approaches based on the re-projection loss \cite{monodepth_unsupervised1,monodepth_unsupervised2} using the backward-warping \cite{stn} have been proposed for monocular depth estimation \cite{monodepth_unsupervised1,monodepth_unsupervised2} or stereo matching \cite{stereo_unsupervised4}. 
It is known that minimizing the re-projection loss in occluded areas or reflective surfaces, however, brings a negative influence to a stereo matching network \cite{monodepth_unsupervised2,stereo_unsupervised5}. 
Another approach is pixel-level domain adaptation based on image-to-image translation between synthetic and real domains \cite{monodepthda1,monodepthda2,monodepthda3,monodepthda4}. 
In this approach, typical methods translate synthetic images to realistic images and a stereo matching network is trained with translated realistic images and synthetic ground truth disparity maps in a supervised manner. 
However, it is non-trivial to generalize this approach to stereo matching, because the image-to-image translation is usually applied to each view separately and the geometric structure of a stereo image pair is not preserved. 
This causes inconsistency between the translated image and the synthetic ground truth disparity map, and training with these noisy labels no longer provides an effective guide signal to the stereo matching network. 
This problem can be observed in Figure \ref{fig:translation_results}, where the image-to-image translation applied to each view separately causes stereo-inconsistency (red rectangles) even in non-occluded areas.
To address this problem, in this paper, we propose an attention mechanism that aggregates features in the left and right views, which is called Stereoscopic Cross Attention (SCA). 
Incorporating SCA to an image-to-image translation network makes it possible to preserve the geometric structure of a stereo image pair in the process of image-to-image translation. 
As can be seen from Figure \ref{fig:translation_results}, the stereo image pair translated with SCA preserves stereo-consistency compared with that translated separately. 
Our contributions are summarized as follows:
\begin{itemize}
  \item[\textbullet] We propose a feature aggregation mechanism called SCA to preserve the geometric structure of a stereo image pair in the process of image-to-image translation.
  \item[\textbullet] We extend the existing image-to-image translation network with SCA, and propose loss functions to guide the image-to-image translation network.
  \item[\textbullet] We empirically demonstrate the effectiveness of the proposed unsupervised domain adaptation based on the image-to-image translation with SCA.
\end{itemize}

\section{Related Work}

\subsection{Unsupervised domain adaptation}

A typical unsupervised domain adaptation approach is making feature representations domain-invariant based on Maximum Mean Discrepancy
(MMD) \cite{mmd1,mmd2}, Wasserstein metric \cite{jdot,deepjdot}, or adversarial learning \cite{dann,confusion,adda}, where an estimator trained on a source domain using domain-invariant features performs well on the target domain. 
Another approach is learning a mapping from a source domain to a target domain, which can be regarded as pixel-level domain adaptation \cite{pixelda,simgan} based on image-to-image translation, where an estimator is trained with the mapped source images and the source labels. This approach has been applied to monocular depth estimation \cite{monodepthda1,monodepthda2,monodepthda3,monodepthda4}, but it is non-trivial to generalize this approach to stereo matching, because the image-to-image translation is usually applied to each view separately, and the geometric structure of a stereo image pair is not preserved. 
Some works proposed style transfers for a stereo image pair \cite{stereo_style_transfer1,stereo_style_transfer2}, but these approaches have not been applied to domain adaptation for stereo matching. 
Gong \textit{et al.} \cite{stereo_style_transfer1} proposed a feature aggregation mechanism using the backward-warping \cite{stn} to generate a stylized stereo image pair preserving stereo-consistency. 
Chen \textit{et al.} \cite{stereo_style_transfer2} also proposed a stereo consistent style transfer, where features in the left and right views are combined in an intermediate middle domain using forward-warping to preserve the stereo-consistency. 
Recently, Liu \textit{et al.} \cite{stereogan} proposed a joint training framework for domain adaptation and stereo matching using loss functions for stereo-consistency, but the image-to-image translation is still applied to each view separately. 
Unlike \cite{stereogan}, we explicitly build paths to share the information between the left and right views and minimize loss functions for stereo-consistency.
We show this mechanism provides a more effective guide signal to the stereo matching network.

\subsection{Deep stereo matching}

Stereo matching has been widely investigated for several decades and a vast number of approaches based on local \cite{adaptive_support}, global \cite{graph_cut} or semi-global \cite{sgm} optimization have been proposed. 
Recently DNN-based stereo matching methods have achieved great progress. 
Mayer \textit{et al.} \cite{scene_flow} proposed the first end-to-end stereo matching network \textit{DispNet} which is trained in a supervised manner to regress a dense disparity map directly. 
Followed by \textit{DispNet}, a lot of end-to-end stereo architectures have been proposed, employing 3D convolution \cite{gcnet,psmnet}, group-wise correlation \cite{gwcnet}, and anisotropic diffusion \cite{cspn}. 
These supervised approaches, however, require a large number of ground truth disparity maps and collecting them in each deployment environment is cumbersome and costly because it requires an expensive depth sensor and careful calibration between the depth sensor and the stereo cameras. 
For this reason, some works proposed frameworks where a stereo matching network is trained in an unsupervised manner, employing a traditional stereo matching algorithm \cite{stereo_unsupervised2}, solving an iterative optimization problem with graph Laplacian regularization \cite{stereo_unsupervised3} or minimizing the re-projection loss \cite{monodepth_unsupervised1,monodepth_unsupervised2,stereo_unsupervised4} using the backward-warping \cite{stn}.
The re-projection loss is defined as the distance (\textit{e.g.} $L_{1}$) between an image in one view and the backward-warped image in the other view.
Tonioni \textit{et al.} \cite{stereo_unsupervised4} proposed an unsupervised online adaptation method for stereo matching based on the re-projection loss. 
It is known that minimizing the re-projection loss in occluded areas or reflective surfaces, however, brings a negative influence to the stereo matching network \cite{monodepth_unsupervised2,stereo_unsupervised5}.
So in \cite{stereo_unsupervised5}, they proposed to learn a confidence estimator for masking the re-projection loss in a meta-learning framework.
The recent trend in stereo matching is to train stereo networks using proxy disparity labels.
\cite{stereo_from_mono} proposed a method to synthesize a stereo image pair which is consistent with a disparity map predicted by a pre-trained monocular depth estimation network.
\cite{monocular_distillation} employs a monocular completion network to generate proxy disparity labels.
Recently, as a method that does not require additional fine-tuning or adaptation, DSMNet \cite{dsmnet} was proposed, where a novel domain normalization layer to reduce the domain shift is designed.

\section{Method}

Let $\mathcal{D}_{s}$ be a source dataset consisting of $N$ tuples $\{(I_{s}^{l},I_{s}^{r},D_{s}^{l},D_{s}^{r})_{i}\}_{i=1}^{N}$ sampled from a source domain and Let $\mathcal{D}_{t}$ be a target dataset consisting of $M$ tuples $\{(I_{t}^{l},I_{t}^{r})_{i}\}_{i=1}^{M}$ sampled from a target domain, where $I_{s}^{l},I_{t}^{l}$ and $I_{s}^{r},I_{t}^{r}$ denotes a left and right image respectively, and $D_{s}^{l}$ and $D_{s}^{r}$ denotes a left and right ground truth disparity map respectively.
Our goal is to learn a stereo matching network that performs well on the target domain.
The basic flow of the proposed framework is as follows. 
First, an image-to-image translation network $\mathcal{G}$ translates stereo image pairs on the source domain to target-style stereo image pairs preserving the geometric structure of them.
Then, a stereo matching network $\mathcal{E}$ is trained with target-style stereo image pairs translated from source stereo image pairs and source ground truth disparity maps. A domain discriminator $\mathcal{C}$ on the target domain is also trained for adversarial learning \cite{gan}. 
In the following sections, we explain the proposed Stereoscopic Cross Attention (SCA) mechanism (\ref{sec:sca}) and loss functions to guide each component in our framework (\ref{sec:loss_adv}, \ref{sec:loss_stereo}, \ref{sec:loss_disp}, \ref{sec:loss_reproj}, \ref{sec:full_objective}).

\begin{figure*}[t]
    \centering
        \includegraphics[width=18cm]{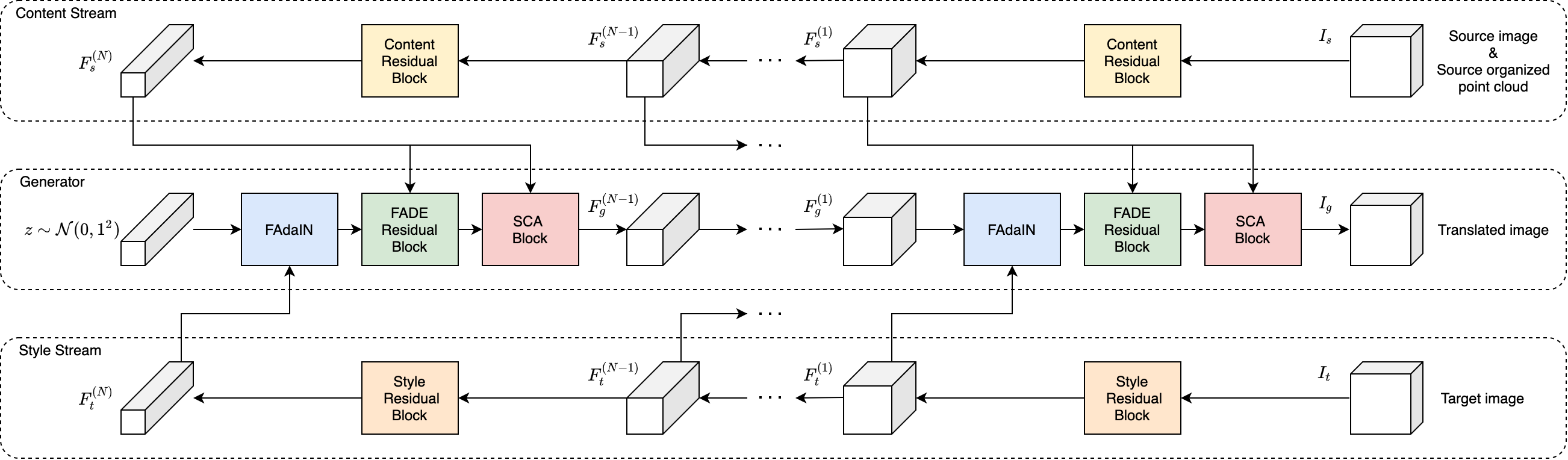}
        \caption{The architecture of the TSIT network \cite{tsit} extended by SCA. The content/style residual block, FAdaIN module and FADE residual block are the ones proposed in \cite{tsit}. $(F_{s})^{n},(F_{t})^{n},(F_{g})^{n}$ denotes the intermediate feature map at the $n$th layer of the content stream, style stream, and generator respectively. $z$ is a latent code sampled from a Gaussian distribution.}
        \label{fig:translation_network}
\end{figure*}

\begin{figure*}[t]
    \centering
        \includegraphics[width=18cm]{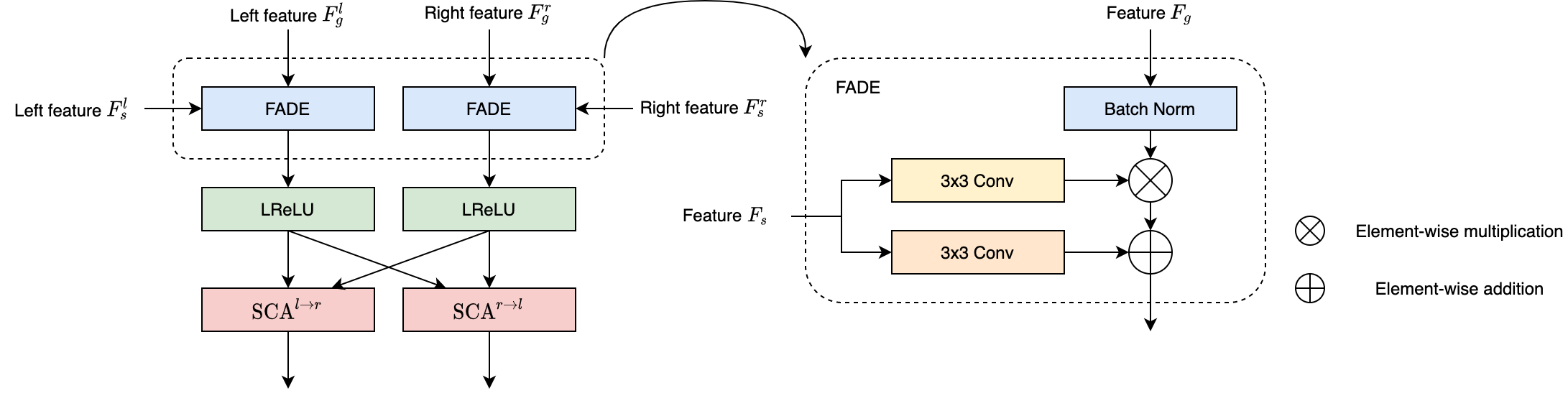}
        \caption{The architecture of the proposed SCA block. The FADE module is the one proposed in \cite{tsit}.}
        \label{fig:sca_block}
\end{figure*}

\subsection{Stereoscopic Cross Attention (SCA)}
\label{sec:sca}

Existing image-to-image translation methods are not designed for a stereo image pair and its geometric structure is not preserved if the image-to-image translation is applied to each view separately.
This means the appearance of a point in one view and that of the corresponding point in the other view can be very different after the image-to-image translation.
This causes inconsistency between the translated stereo image pair and the source ground truth disparity map, and training with these noisy labels no longer provides an effective guide signal to the stereo matching network. 
To address this problem, we propose an attention mechanism that aggregates features in the left and right views, called Stereoscopic Cross Attention (SCA). 
By incorporating SCA to the image-to-image translation network $\mathcal{G}$, we can explicitly build paths to share the information between the left and right views to preserve the geometric structure of a stereo image pair. 
To achieve the above mechanism, we introduce Attention \cite{attention,transformer}. 
Attention is a mechanism to compute a summation of \textit{value} weighted by similarity of \textit{query} and \textit{key}.
For the proposed SCA, \textit{query} is a point-wise feature in one view, \textit{key} and \textit{value} are point-wise features on the epipolar line in the other view. 
By this mechanism, SCA learns "\textit{Which point should I attend to preserve the stereo-consistency?}".
In order to let SCA to find the corresponding point in the other view easily, we input the organized 3D point cloud in a world coordinate system re-projected from a ground truth disparity map to the image-to-image translation network $\mathcal{G}$ in addition to a RGB image.
Given a dense disparity map $D^{v}$, where $v \in \{l,r\}$ denotes the left or right view, the organized 3D point cloud in the camera coordinate system of the view $v$ denoted as $P_{cam}^{v}=\{(x^{v}_{ij},y^{v}_{ij},z^{v}_{ij})\}_{ij}$ is calculated as follows:
\begin{align}
    x^{v}_{ij}=&\cfrac{b(u-c_{u})}{D^{v}_{ij}}\ , \\
    y^{v}_{ij}=&\cfrac{f_{u}b(v-c_{v})}{f_{v}D^{v}_{ij}}\ , \\
    z^{v}_{ij}=&\cfrac{f_{u}b}{D^{v}_{ij}}\ ,
\end{align}
where $b$ denotes the baseline between the stereo cameras, $(c_{u},c_{v})$ denotes the pixel location corresponding to the
camera center, and $f_{u},f_{v}$ denotes the horizontal and vertical focal length.
In order to represent $P_{cam}^{l},P_{cam}^{r}$ in a world coordinate system, we take the middle between the stereo camera centers as the origin, so the organized 3D point cloud in the world coordinate system is represented as follows:
\begin{align}
    P^{l}=\{(x^{l}_{ij}-\cfrac{b}{2},y^{l}_{ij},z^{l}_{ij})\}_{ij}\ , \\
    P^{r}=\{(x^{r}_{ij}+\cfrac{b}{2},y^{r}_{ij},z^{r}_{ij})\}_{ij}\ .
\end{align}
Therefore, given a feature map $F^{v} \in \mathbb{R}^{d_{in} \times H \times W}$ extracted from a RGB image $I^{v}$ and an organized point cloud $P^{v}$, and learned matrices $W_{Q},W_{K} \in \mathbb{R}^{d_{out} \times 2d_{in}}$, where $v \in \{l,r\}$ denotes the left or right view, SCA from the left view to the right view denoted as $\text{SCA}^{l \rightarrow r}$ and that from the right view to the left view denoted as $\text{SCA}^{r \rightarrow l}$ are formulated as follows:
\begin{align}
    \text{SCA}^{l \rightarrow r}_{ij}&=\sum_{d=0}^{d_{max}}\softmax_{d}((Q^{l}_{ij})^{T}K^{r}_{i-d,j})F^{r}_{i-d,j}\ , \\
    \text{SCA}^{r \rightarrow l}_{ij}&=\sum_{d=0}^{d_{max}}\softmax_{d}((Q^{r}_{ij})^{T}K^{l}_{i+d,j})F^{l}_{i+d,j}\ ,
\end{align}
where \textit{query} $Q^{v}=W_{Q}F^{v}$, \textit{key} $K^{v}=W_{K}F^{v}$, and $d_{max}$ is the maximum disparity that the stereo matching network considers.
We incorporate SCA to the image-to-image translation network $\mathcal{G}$.
In this paper, the translation network $\mathcal{G}$ is based on the one used in TSIT \cite{tsit}, a recently proposed image-to-image translation framework. 
We extended TSIT by the proposed SCA as shown in Figure \ref{fig:translation_network}.
TSIT consists of two feature extractors and one generator.
One feature extractor is \textit{Content Stream} for a source domain and the other is \textit{Style Stream} for a target domain. Multi-scale feature representations learned by the content/style stream is fed to the feature transformation layers called \textit{FAdaIN} and \textit{FADE Residual Block} in the generator.
The generator takes a latent code sampled from Gaussian distribution as the input. At each scale, the FAdaIN module fuses a style representation of a target image learned by the style stream and the FADE residual block fuses a content representation of a source image learned by the content stream.
We implement SCA as a SCA block inspired by the FADE residual block, but designed to preserve stereo-consistency unlike the FADE residual block.
The architecture of the SCA block is shown in Figure \ref{fig:sca_block}.
The SCA block is followed by the FADE residual block and preserves stereo-consistency using the feature extracted from the content stream at each scale.

\subsection{Adversarial loss}
\label{sec:loss_adv}

In our method, we train the translation network $\mathcal{G}$ to learn a mapping from the source domain to the target domain by adversarial learning \cite{gan} using a domain discriminator $\mathcal{C}$ on the target domain.
$\mathcal{G}$ and $\mathcal{C}$ are trained in an alternating fashion by minimizing the following hinge version of the adversarial loss functions \cite{geogan,spectral_norm}:

\normalsize
\begin{align}
    &\mathcal{L}^{\mathcal{G}}_{adv}(\mathcal{G},\mathcal{D}_{s})
    =-\mathbb{E}_{(I_{s}^{l},I_{s}^{r},D_{s}^{l},D_{s}^{r}) \sim \mathcal{D}_{s},(I_{t}^{l},I_{t}^{r}) \sim \mathcal{D}_{t}}\nonumber \\
    &[\sum_{\substack{(b,m) \in \\ \{(l,r),(r,l)\}}}\mathcal{C}(\mathcal{G}(I_{s}^{b},I_{s}^{m},D_{s}^{b},D_{s}^{m},I_{t}^{b},I_{t}^{m}))]\ , \nonumber \\ \\
    &\mathcal{L}^{\mathcal{C}}_{adv}(\mathcal{C},\mathcal{D}_{s},\mathcal{D}_{t})=\mathbb{E}_{(I_{s}^{l},I_{s}^{r},D_{s}^{l},D_{s}^{r}) \sim \mathcal{D}_{s},(I_{t}^{l},I_{t}^{r}) \sim \mathcal{D}_{t}} \nonumber \\
    &[\sum_{\substack{(b,m) \in \\ \{(l,r),(r,l)\}}}\max(0,1+\mathcal{C}(\mathcal{G}(I_{s}^{b},I_{s}^{m},D_{s}^{b},D_{s}^{m},I_{t}^{b},I_{t}^{m})))] \nonumber \\
    +&\mathbb{E}_{(I_{s}^{l},I_{s}^{r}) \sim \mathcal{D}_{s}}[\sum_{v \in \{l,r\}}\max(0,1+\mathcal{C}(I_{s}^{v}))] \nonumber \\
    +&\mathbb{E}_{(I_{t}^{l},I_{t}^{r}) \sim \mathcal{D}_{t}}[\sum_{v \in \{l,r\}}\max(0,1-\mathcal{C}(I_{t}^{v}))]\ ,
\end{align}
\normalsize

where $\mathcal{G}(I_{s}^{b},I_{s}^{m},P_{s}^{b},P_{s}^{m},I_{t}^{b},I_{t}^{m})$ denotes the translated image in a view $b$.

Moreover, for the translation network $\mathcal{G}$, we apply the perceptual loss $\mathcal{L}_{perc}(\mathcal{G},\mathcal{D}_{s},\mathcal{D}_{t})$ \cite{perceptual} and the feature matching loss $\mathcal{L}_{feat}(\mathcal{G},\mathcal{D}_{s},\mathcal{D}_{t})$ \cite{pix2pixhd} following \cite{tsit}.

\begin{table*}[tb]
  \caption{The domain adaptation results with and without using SCA.}
  \label{tab:ablation_study}
  \centering
    \scalebox{1.4}{
        \begin{tabular}{c|c|c|c|c|c} \hline
        Source dataset           & Target dataset                     & Method         & Backbone                                    & D1-all [\%]   & EPE [px] \\ \hline
        \multirow{2}{*}{Driving} & \multirow{2}{*}{KITTI Stereo 2015} & w/o SCA        & \multirow{2}{*}{DispNetC \cite{scene_flow}} & 10.11         & 1.64 \\
                                 &                                    &\textbf{w/ SCA} &                                             & \textbf{8.40} & \textbf{1.47} \\ \hline
        \end{tabular}
    }
\end{table*}

\begin{table*}[tb]
  \caption{The evaluation results of the proposed method compared to other methods.}
  \label{tab:adaptation_results}
  \centering
    \scalebox{1.4}{
        \begin{tabular}{c|c|c|c|c|c} \hline
            Source dataset                   & Target dataset                     & Method                              & Backbone                                    & D1-all [\%]   & EPE [px] \\ \hline
            \multirow{5}{*}{Driving}         & \multirow{1}{*}{None}              & Inference                           & \multirow{5}{*}{DispNetC \cite{scene_flow}} & 85.94         & 21.22  \\ \cline{2-2}
                                             & \multirow{4}{*}{KITTI Stereo 2015} & SL+Ad \cite{stereo_unsupervised4}   &                                             & 39.49         & 4.77 \\
                                             &                                    & L2A+Wad \cite{stereo_unsupervised5} &                                             & 26.90         & 3.01 \\
                                             &                                    & StereoGAN \cite{stereogan}          &                                             & 25.71         & 2.75 \\
                                             &                                    &\textbf{Proposed}                    &                                             & \textbf{8.40} & \textbf{1.47} \\ \cline{1-6}
            \multirow{3}{*}{Scene Flow}      & \multirow{2}{*}{None}              & Inference                           & GwcNet \cite{gwcnet}                        & 31.58         & 7.86  \\
                                             &                                    & DSMNet \cite{dsmnet}                & DSMNet \cite{dsmnet}                        & 6.50          & - \\ \cline{2-2}
                                             & \multirow{1}{*}{KITTI Stereo 2015} &\textbf{Proposed}                    & GwcNet \cite{gwcnet}                        & \textbf{5.19} & \textbf{1.12} \\ \cline{1-6}
        \end{tabular}
    }
\end{table*}

\subsection{Stereo-consistency loss}
\label{sec:loss_stereo}

By incorporating SCA, the translation network $\mathcal{G}$ now has a capacity to aggregate features in the left and right views.
In order to guide SCA to preserve stereo-consistency, we introduce stereo-consistency loss based on the re-projection loss \cite{monodepth_unsupervised2}.
We do not use a SSIM term \cite{ssim} unlike in \cite{monodepth_unsupervised2}, so the stereo-consistency loss is just defined as $L_{1}$ loss between an image in one view and the backward-warped image in the other view.
We do not minimize this stereo-consistency loss in occluded areas.
Moreover, we apply this stereo-consistency loss on the multi-scale feature maps from the translation network $\mathcal{G}$.
This stereo-consistency loss is formulated as follows:

\footnotesize
\begin{align}
    &\mathcal{L}_{stereo}(\mathcal{G},D_{s},D_{t})
    =\mathbb{E}_{(I_{s}^{l},I_{s}^{r},D_{s}^{l},D_{s}^{r}) \sim \mathcal{D}_{s},(I_{t}^{l},I_{t}^{r}) \sim \mathcal{D}_{t}} \nonumber \\
    &[\sum_{\substack{(b,m) \in \\ \{(l,r),(r,l)\}}}\sum_{n}^{N}\cfrac{\sum_{ij}\|(F_{g}^{b})_{ij}^{(n)}-\overset{\leftarrow}{W}((F_{g}^{m})^{(n)}, D_{s}^{b})_{ij}\|_{1} \odot (M_{s}^{b})_{ij}}{\sum_{ij} (M_{s}^{b})_{ij}}]\ ,
\end{align}
\normalsize

where $(F_{g}^{b})^{(n)}=\uparrow_{n}(\mathcal{G}^{(n)}(I_{s}^{b},I_{s}^{m},D_{s}^{b},D_{s}^{m},I_{t}^{b},I_{t}^{m}))$ and $(F_{g}^{m})^{(n)}=\uparrow_{n}(\mathcal{G}^{(n)}(I_{s}^{m},I_{s}^{b},D_{s}^{m},D_{s}^{b},I_{t}^{m},I_{t}^{b}))$ denotes the up-sampled feature map at the $n$th layer of the translation network $\mathcal{G}$ in a view $b$ and $m$ respectively, $\uparrow_{n}(\cdot)$ denotes a $2^{n}$ times up-sampling operator, $\overset{\leftarrow}{W}(\cdot,\cdot)$ denotes the backward-warping function \cite{stn}, and $M_{s}^{b}$ denotes the occlusion mask in a view $b$.

The backward-warping function $\overset{\leftarrow}{W}(\cdot,\cdot)$ takes a feature map in a view $m$ and a dense disparity map from a view $b$ to the view $m$ as inputs, and outputs the backward-warped feature map in the view $b$ as follows:
\begin{align}
    \overset{\leftarrow}{W}(F^{m},D^{b})_{ij}=&\sum_{k}\max(0,1-|i+D^{b}_{ij}-k|)F^{m}_{kj}\ .
\end{align}
The occlusion mask $M_{s}^{b}$ is calculated using so called left-right consistency check as follows:
\begin{align}
    M_{s}^{b}=\mathds{1}(|D_{s}^{b}-\overset{\leftarrow}{W}(D_{s}^{m},D_{s}^{b})|<1)\ ,
\end{align}
where $\mathds{1}(\cdot)$ denotes an indicator function.

\subsection{Disparity loss}
\label{sec:loss_disp}

By minimizing $\mathcal{L}_{adv}$, translated images by $\mathcal{G}$ get closer to those on the target domain.
Therefore, we can get an optimal stereo matching network on the target domain by supervised training with translated stereo image pairs and source ground truth disparity maps.
We note that this supervised training provides an effective guide signal only when each translated stereo image pair is consistent with the source ground truth disparity map.
This supervised loss is formulated as follows:
\begin{align}
    \mathcal{L}_{disp}(\mathcal{G},\mathcal{E},\mathcal{D}_{s},\mathcal{D}_{t})=\mathbb{E}_{(I_{s}^{l},I_{s}^{r},D_{s}^{l},D_{s}^{r}) \sim \mathcal{D}_{s}, (I_{t}^{l},I_{t}^{r}) \sim \mathcal{D}_{t}} \nonumber \\
    [\sum_{(b,m) \in \{(l,r),(r,l)\}}\text{Smooth}_{L_{1}}(\hat{D}_{g}^{b}-D_{s}^{b})]\ ,
\end{align}
where $\hat{D}_{g}^{b}=\mathcal{E}(I_{g}^{b},I_{g}^{m}))$ denotes the predicted disparity map, $I_{g}^{b}=\mathcal{G}(I_{s}^{b},I_{s}^{m},D_{s}^{b},D_{s}^{m},I_{t}^{b},I_{t}^{m})$ and $I_{g}^{m}=\mathcal{G}(I_{s}^{m},I_{s}^{b},D_{s}^{m},D_{s}^{b},I_{t}^{m},I_{t}^{b})$ denotes the translated image in a view $b$ and $m$ respectively.
The $\text{Smooth}_{L_{1}}$ loss is defined as follows:
\begin{align}
    \text{Smooth}_{L_{1}}(x)=
    \begin{cases}
    0.5x^{2} & \text{if } |x| < 1 \\
    |x|-0.5 & \text{otherwise}\ .
    \end{cases}
\end{align}

\subsection{Re-projection loss}
\label{sec:loss_reproj}

The re-projection loss \cite{monodepth_unsupervised2} can also be used as an unsupervised loss for stereo matching network $\mathcal{E}$ on the target domain. 
The re-projection loss is defined as a combination of SSIM \cite{ssim} and $L_{1}$.
While the stereo-consistency loss $\mathcal{L}_{stereo}$ for the translation network $\mathcal{G}$ guides each translated stereo image pair to be consistent with the source ground truth disparity map, the re-projection loss for the stereo matching network $\mathcal{E}$ guides the predicted disparity map to be consistent with the target stereo image pair.
The re-projection loss is formulated as follows:

\footnotesize
\begin{align}
    &\mathcal{L}_{reproj}(\mathcal{E},\mathcal{D}_{t})
    =\mathbb{E}_{(I_{t}^{l},I_{t}^{r}) \sim \mathcal{D}_{t}}
    [\sum_{(b,m) \in \{(l,r),(r,l)\}} \nonumber \\
    &(1-\alpha)\|I_{t}^{b}-\overset{\leftarrow}{W}(I_{t}^{m},\hat{D}_{t}^{b})\|_{1}
    +\cfrac{\alpha}{2}(1-\text{SSIM}(I_{t}^{b},\overset{\leftarrow}{W}(I_{t}^{m},\hat{D}_{t}^{b})))]\ ,
\end{align}
\normalsize
where the predicted disparity $\hat{D}_{t}^{b}=\mathcal{E}(I_{t}^{b},I_{t}^{m})$, and $\alpha=0.85$.

\subsection{Full objective}
\label{sec:full_objective}

From the above, we can get an optimal stereo matching network $\mathcal{E}^{*}$ on the target domain by solving the following optimization problem:
\begin{align}
    \mathcal{E}^{*}=&\argmin_{\mathcal{E}}\min_{\mathcal{G},\mathcal{C}}
    \mathcal{L}^{\mathcal{G}}_{adv}(\mathcal{G},\mathcal{D}_{s},\mathcal{D}_{t})
    +\mathcal{L}^{\mathcal{C}}_{adv}(\mathcal{C},\mathcal{D}_{s},\mathcal{D}_{t})\nonumber \\
    +&\lambda_{perc}\mathcal{L}_{perc}(\mathcal{G},\mathcal{D}_{s},\mathcal{D}_{t})
    +\lambda_{feat}\mathcal{L}_{feat}(\mathcal{G},\mathcal{D}_{s},\mathcal{D}_{t})\nonumber \\
    +&\lambda_{stereo}\mathcal{L}_{stereo}(\mathcal{G},\mathcal{D}_{s},\mathcal{D}_{t})
    +\lambda_{disp}\mathcal{L}_{disp}(\mathcal{G},\mathcal{E},\mathcal{D}_{s},\mathcal{D}_{t})\nonumber \\
    +&\lambda_{reproj}\mathcal{L}_{reproj}(\mathcal{E},\mathcal{D}_{t})\ ,
    \label{eq:full_objective}
\end{align}
where $\lambda_{perc}, \lambda_{feat}, \lambda_{stereo}, \lambda_{disp}, \lambda_{reproj}$ are hyper parameters that weight the corresponding loss.

\begin{figure*}[t]
    \centering
        \begin{tabular}{cc}
          \begin{minipage}{0.45\hsize}
            \centering
            \includegraphics[width=8cm]{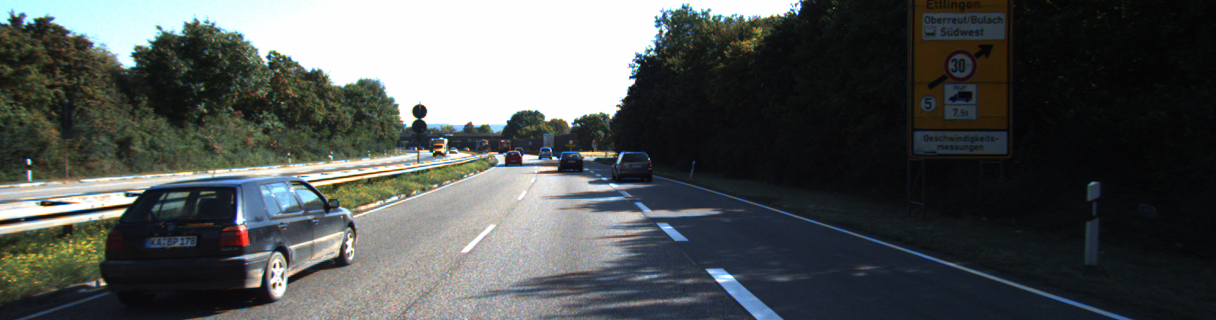}
            \subcaption{Left input image}
          \end{minipage} &
          \begin{minipage}{0.45\hsize}
            \centering
            \includegraphics[width=8cm]{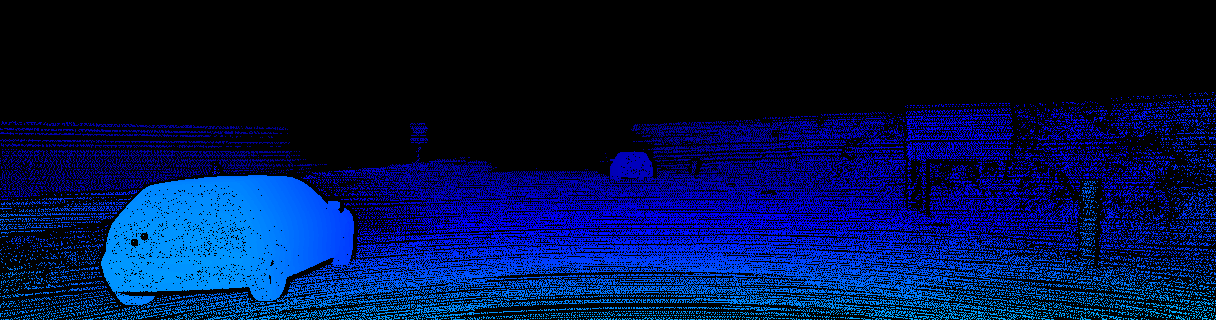}
            \subcaption{Ground truth disparity map}
          \end{minipage}\\
          \begin{minipage}{0.45\hsize}
            \centering
            \includegraphics[width=8cm]{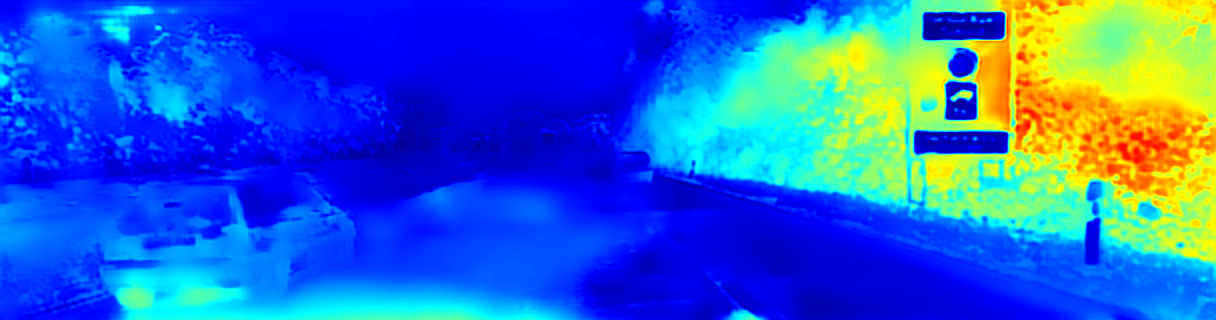}
            \subcaption{Predicted disparity map before adaptation}
          \end{minipage} &
          \begin{minipage}{0.45\hsize}
            \centering
            \includegraphics[width=8cm]{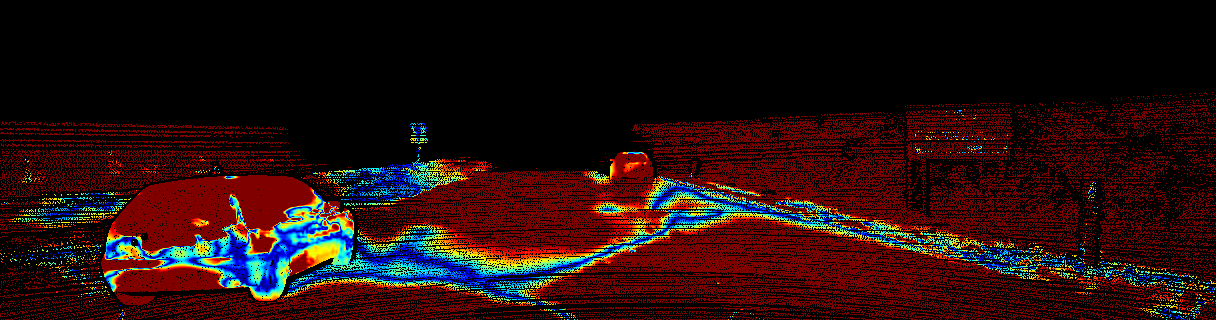}
            \subcaption{Error map before adaptation}
          \end{minipage}\\
          \begin{minipage}{0.45\hsize}
            \centering
            \includegraphics[width=8cm]{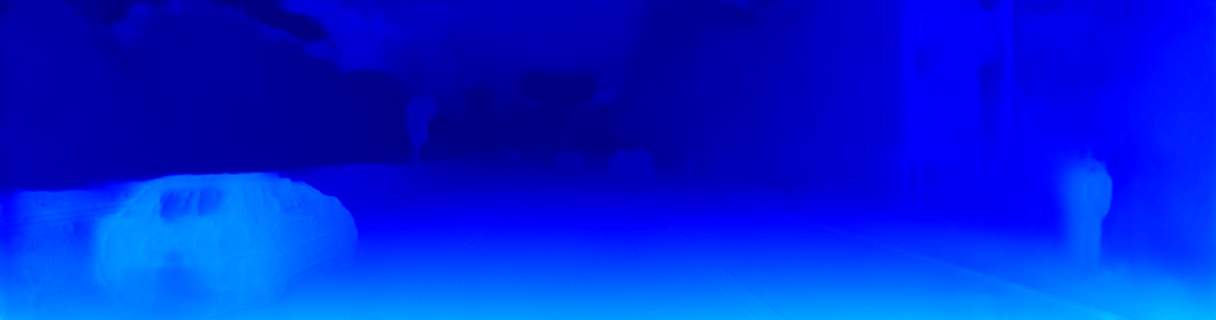}
            \subcaption{Predicted disparity map after adaptation}
          \end{minipage} &
          \begin{minipage}{0.45\hsize}
            \centering
            \includegraphics[width=8cm]{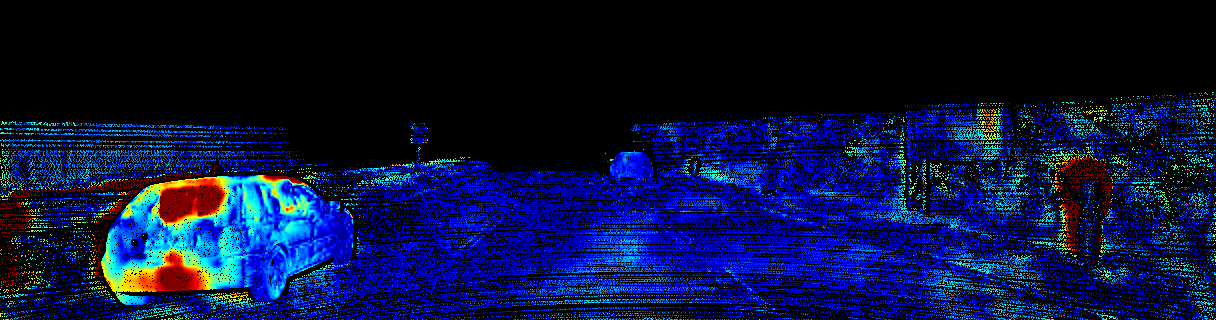}
            \subcaption{Error map after adaptation}
          \end{minipage}
        \end{tabular}
        \caption{The stereo matching results on KITTI Stereo 2015 before and after applying the proposed domain adaptation from Driving to KITTI Stereo 2015. Blue pixels indicate smaller values and red pixels indicate larger values in the error maps. We scale the error maps so that 3 px error corresponds the intermediate color, green.}
        \label{fig:adaptation_results}
\end{figure*}

\section{Experiments}

\subsection{Datasets}

In this paper, we use the synthetic datasets, Scene Flow \cite{scene_flow} and Driving which is a subset of Scene Flow as our source datasets, and the real dataset,  KITTI Stereo 2015 \cite{kitti2015} as our target dataset. 
For KITTI Stereo 2015, we use KITTI Stereo 2012 \cite{kitti2012} training set as the training and validation set, and KITTI Stereo 2015 training set as the test set. 
Moreover, we use KITTI 3D object 2017 \cite{kitti2012} as the additional training set for all unsupervised methods.

\subsection{Evaluation metrics}

We use both end-point error (EPE) and D1-all as the evaluation metrics.
EPE is the disparity error averaged in all the pixels.
D1-all is the percentage of disparity outliers in all the pixels. 
The outliers are defined as the pixels whose disparity error is larger than
$\text{max}(3, 0.05d)$, where $d$ denotes the ground truth disparity.

\subsection{Implementation details}

We use DispNetC \cite{scene_flow} and GwcNet-g \cite{gwcnet} as the stereo matching network $\mathcal{E}$. 
We use multi-scale 70 × 70 PatchGAN \cite{pix2pixhd} as the domain discriminator $\mathcal{C}$.
We apply spectral normalization \cite{spectral_norm} to all the convolution kernels in the domain discriminator $\mathcal{C}$.
We first train the stereo matching network $\mathcal{E}$ on the source dataset with $L_{1}$ loss in a supervised manner, using Adam optimizer \cite{adam} with the momentum $\beta_{1}=0.9$, $\beta_{2}=0.999$, and the learning rate $\alpha=0.0001$ with a batch size of $32$.
Then, we train the image-to-image translation network $\mathcal{G}$ and the domain discriminator $\mathcal{C}$ with $\mathcal{L}^{\mathcal{G}}_{adv}, \mathcal{L}^{\mathcal{C}}_{adv}, \mathcal{L}_{perc}, \mathcal{L}_{feat}, \mathcal{L}_{stereo}$, using Adam optimizer with the momentum $\beta_{1}=0.0$, $\beta_{2}=0.9$, and the learning rate $\alpha=0.0001$ for $\mathcal{G}$ and $\alpha=0.0004$ for $\mathcal{C}$ with a batch size of $32$. 
Finally, we train the stereo matching network $\mathcal{E}$ with $\mathcal{L}^{disp}$ and $\mathcal{L}^{reproj}$, using Adam optimizer with the momentum $\beta_{1}=0.9$, $\beta_{2}=0.999$, and the learning rate $\alpha=0.0001$ with a batch size of $32$.
We set the weighting hyper parameters as $\lambda_{perc}=1.0, \lambda_{feat}=1.0, \lambda_{stereo}=10.0, \lambda_{disp}=0.1, \lambda_{reproj}=1.0$.

\subsection{Ablation study}

To confirm the effectiveness of the proposed SCA, we compare the domain adaptation results between with and without SCA.
We take Driving as our source dataset, KITTI Stereo 2015 as our target dataset, and DispNetC \cite{scene_flow} as our stereo matching network $\mathcal{E}$.
The results are shown in Table \ref{tab:ablation_study}.
As can be seen from Table \ref{tab:ablation_study}, the domain adaptation with SCA-based feature aggregation outperforms that without any feature aggregation.
This shows that preserving the stereo-consistency by SCA in the process of the image-to-image translation provides a more effective guide signal to the stereo matching network.

\subsection{Quantitative evaluation}

We compare the proposed method with state-of-the-art unsupervised methods.
We take Driving and Scene Flow as our source dataset, KITTI Stereo 2015 as our target dataset, and DispNetC \cite{scene_flow} and GwcNet \cite{gwcnet} as our stereo matching network $\mathcal{E}$.
The results are shown in Table \ref{tab:adaptation_results}.
\textit{Inference} denotes our stereo matching network $\mathcal{E}$ trained only on the source dataset with $L_{1}$ loss and directly evaluated on the target dataset without any fine-tuning.
SL+Ad denotes an unsupervised online adaptation method proposed in \cite{stereo_unsupervised4} and L2A+Wad denotes an unsupervised adaptation method via meta learning framework proposed in \cite{stereo_unsupervised5}.
StereoGAN \cite{stereogan} is a recently proposed joint training framework for domain adaptation and stereo matching.
We note that GwcNet \cite{gwcnet} has fewer parameters parameters than DSMNet \cite{dsmnet}.
So there is no accuracy improvement by the number of parameters of the backbone network in comparison with DSMNet \cite{dsmnet}.
As can be seen from Table \ref{tab:adaptation_results}, the proposed method surpasses all the other unsupervised methods on both Driving to KITTI Stereo 2015 and SceneFlow to KITTI Stereo 2015.

\subsection{Qualitative evaluation}

For quantitative evaluation, we show the stereo matching results before and after applying the proposed domain adaptation from Driving to KITTI Stereo 2015 in Figure \ref{fig:adaptation_results}.
As seen in Figure \ref{fig:adaptation_results}, the stereo matching network fails to predict the disparities in most of the regions, especially the trees and the traffic sign, before the domain adaptation, but predicts those correctly after the domain adaptation.

\section{Conclusion}

In this paper, we proposed a feature aggregation mechanism called SCA to preserve the geometric structure of a stereo image pair in the process of image-to-image translation.
We extended the existing image-to-image translation network with SCA and empirically demonstrated the effectiveness of the proposed unsupervised domain adaptation based on the image-to-image translation with SCA.


\bibliographystyle{IEEEtran}
\bibliography{main}

\end{document}